\documentclass[a4paper,fleqn]{cas-dc}

\usepackage[authoryear]{natbib}

\makeatletter
\renewcommand\hyper@natlinkbreak[2]{#1}
\makeatother

\usepackage{amsmath}
\usepackage{hyperref}

\usepackage{float}
\usepackage{balance}

\def\tsc#1{\csdef{#1}{\textsc{\lowercase{#1}}\xspace}}
\tsc{KT}
\tsc{ITS}
\tsc{PICKT}
\tsc{LiLT}
\tsc{HAN}

\usepackage{caption}
\usepackage{arydshln}  
\usepackage{multirow}  

\usepackage{tabularx}
\usepackage{array}

\usepackage{kotex}
\usepackage{placeins}

\begin{document}
\let\printorcid\relax

\ExplSyntaxOn
\cs_gset:Npn \__first_footerline: { }
\ExplSyntaxOff

\makeatletter
\def\ps@first{%
  \let\@oddhead\@empty
  \let\@evenhead\@empty
  \let\@oddfoot\@empty
  \let\@evenfoot\@empty
}
\def\ps@plain{%
  \let\@oddhead\@empty
  \let\@evenhead\@empty
  \def\@oddfoot{\hfil\thepage\hfil}%
  \let\@evenfoot\@oddfoot
}
\makeatother

\pagestyle{plain}

\let\WriteBookmarks\relax
\def\floatpagepagefraction{1}
\def\textpagefraction{.001}

\shorttitle{}

\shortauthors{}

\title [mode = title]{Enhancing knowledge tracing robustness for new question cold start in Intelligent Tutoring Systems}



%





\author[1]{Wonbeen Lee}
\cormark[1]
\ead{wonbeeny@gmail.com}

\author[1]{Channyung Lee}
\ead{icn971234@gmail.com}

\author[1]{Junho Sohn}
\ead{samsohn1004@gmail.com}

\author[1]{Hansam Cho}
\ead{nokopusa@gmail.com}

\affiliation[1]{organization={CHUNJAE EDUCATION INC. AI-Center},
    addressline={16 Gasan digital 1-ro, AP Tower 19th floor}, 
    city={Geumcheon-gu, Seoul},
    postcode={08591}, 
    country={Republic of Korea}}

\cortext[cor1]{Corresponding author}

\begin{abstract}
Intelligent Tutoring Systems (ITS) provide personalized learning paths by diagnosing learners’ proficiency. Knowledge Tracing (KT) models play a central role in this diagnosis by estimating learners’ evolving knowledge states. However, in real-world ITS services, diagnostic reliability may decrease when newly introduced questions have no prior interaction history. This study aims to empirically identify the key features that support KT model robustness under the question cold start situation. To this end, we designed Practical Integrated Cross-consistent Knowledge Tracing (PICKT), which integrates multiple types of features, and examined which feature contributes to robustness under the question cold start situation. Among these features, we further analyzed difficulty, texts, and relational information derived from the knowledge map, which have been identified as important factors in prior KT research, to examine how each contributes to robustness under the question cold start situation. The results showed that difficulty feature was particularly informative for highly challenging questions, where the likelihood of a correct response rate was remarkably low. Meanwhile, the fused texts and knowledge map features supported robustness by enabling the model to estimate the representation of unseen questions by leveraging semantically and structurally similar questions observed during training. These findings suggest that maintaining KT robustness in ITS applications requires prioritization of feature annotation aligned with the characteristics of educational services.
\end{abstract}



\begin{keywords}
Knowledge Tracing \sep Intelligent Tutoring Systems \sep Personalized Learning \sep Cold Start \sep Deep Learning
\end{keywords}

\maketitle

\section{Introduction}\label{sec:1.Introduction}
Intelligent Tutoring Systems (ITS) provide personalized learning paths and customized feedback by accurately diagnosing learners’ knowledge states \citep{alkhatlan2018intelligent, maity2024generative, zhou2025deep}. Accurate diagnosis of learners’ knowledge states is therefore a prerequisite for successful ITS implementation, and Knowledge Tracing (KT) models are the foundation of this diagnostic process \citep{wan2023learning, dai2021knowledge, wang2023multiple}. However, in real-world ITS services, cold start situations frequently arise because new questions are continuously introduced \citep{fu2024sinkt}. Existing KT models may suffer severe performance degradation in these situations because the lack of prior interaction between learners and questions limits their ability to estimate meaningful question representations \citep{zhang2021knowledge, bhattacharjee2025cold}. Therefore, ensuring robustness under the question cold start situations \citep{saveski2014item} is essential for maintaining diagnostic reliability of real-world ITS applications.

To address this data sparsity problem, previous studies have attempted to incorporate features such as question difficulty \citep{liu2024question, kim2023multi}, texts \citep{liu2022open, liu2019ekt, su2018exercise, yin2019quesnet, kim2025knowledge}, and a knowledge map that captures relationships among concepts \citep{wu2023self, nakagawa2019graph, tong2020structure, yang2020gikt}, into KT models. These features can help models to estimate meaningful representations even for questions with no prior interaction \citep{guo2024mitigating}. However, the specific contribution of each feature to model robustness under the question cold start situations has not yet been sufficiently investigated.

In this study, we empirically analyzed the key features that ensure the robustness of KT models under the question cold start scenario. To this end, we propose the Practical Integrated Cross-consistent Knowledge Tracing (PICKT) model, which can integrate various features. PICKT is designed to organically fuse texts and knowledge map features along with the difficulty feature of the questions. In this study, the performance contribution of each feature is closely reviewed using PICKT as an analysis tool.

The results indicate that fused texts and knowledge map, together with question difficulty, contributes substantially to maintaining model robustness under the question cold start scenario. In particular, when the difficulty of the question was very hard, the difficulty feature played an important role in maintaining the predictive performance. Meanwhile, fused texts and knowledge map features were also proven to provide significant information for the model to capture the characteristics of new questions. 

This study suggest that when designing real-world ITS service, the annotation of key features should be prioritized according to the objectives and characteristics of the service to maintain robust predictive performance under the question cold start situation. The difficulty feature should be prioritized in educational services involving a substantial proportion of highly challenging questions, meanwhile the fused texts and knowledge map features should be prioritized in services where questions share semantic or structural similarities. By establishing a stable achievement diagnosis system even under the question cold start situation, this study ultimately aims to contribute to providing learners with a more reliable and personalized learning experience.

\section{Related Work}\label{sec:2.Related Work}

\subsection{Knowledge Tracing}\label{sec:2.1.Knowledge Tracing}

Knowledge Tracing (KT) is a task to estimate the learner's knowledge state based on sequence of interactions and to predict the probability of correct response to questions that the learner has not yet solved \citep{abdelrahman2023knowledge, corbett1994knowledge, bai2024survey}. KT is a key component of the Intelligent Tutoring Systems (ITS), laying the foundation for providing personalized feedback and customized learning experiences to learners. Specifically, based on the knowledge state estimated by the KT model, ITS can recommend appropriate learning materials or follow-up questions. In addition, instructors can use the diagnostic results to add or update learning materials \citep{wan2023learning, shen2024survey}.

\subsection{Cold Start}\label{sec:2.2.Cold Start}

Cold start is a widely known problem in models trained on interaction histories between learners and questions \citep{schein2002methods}. This problem arises because there is little or no past interaction data for new learners or new questions \citep{bhattacharjee2025cold, zhang2021knowledge, bai2025cskt}. The cold start problem caused by new learners \citep{pian2020gamified}, in real-world ITS service can be partially alleviated through initial diagnostic assessments or a predefined curriculum. On the other hand, new questions with no interaction by learners at all are more difficult to handle \citep{guo2024mitigating}. Since a majority of KT models represent questions using ID-based embeddings, their performance is limited \citep{jung2024clst, wang2025llm}. This is because new questions must be represented only by categorical features trained from the interactions of other questions. In real-world online education platforms, new questions are often added \citep{fu2024sinkt} due to new services or changes in the curriculum. Therefore, this question cold start situation is one of the main factors limiting the practical application of the KT model.

To overcome these limitations, the recent KT research has proposed models that incorporate various features of the question \citep{park2024enhancing, li2024mlkt4rec}, and some models have demonstrated robustness under the question cold start situation. Specifically, question difficulty has been used to reduce reliance on question IDs \citep{liu2024question}, while textual information has been incorporated to capture the semantic characteristics of questions \citep{liu2022open}. In addition, graph-based approaches such as graph neural networks, have been actively explored to reflect question–concept relationships and concept-concept relationships \citep{wu2023self, kim2024content, nakagawa2019graph}.

\subsection{Language-independent Layout Transformer}\label{sec:2.3.Language-Independent Layout Transformer}

The Language-independent Layout Transformer (LiLT) \citep{wang2022lilt} is a model used in the field of structured document understanding. In LiLT, the layout encoder layer and text encoder layer operate independently of the language, while exchanging information with each other during the computation process. This interaction structure is designed to reflect interrelationships while maintaining the layout and text independently of their unique characteristics and roles. As a result, LiLT trains different types of data separately. LiLT reflects various features regardless of linguistic characteristics or document structure. Synchronized interaction and independent processing of layout and text are one of the biggest advantages of LiLT, providing flexibility to understand diverse languages and structures \citep{yoon2024language} of documents in practice.

\subsection{Heterogeneous graph Attention Network}\label{sec:2.4.Heterogeneous graph Attention Network}

Heterogeneous graph Attention Network (HAN) \citep{wang2019heterogeneous} is a graph neural network designed to handle heterogeneous graphs including various types of nodes and edges. Heterogeneous graphs include different types of nodes and various relationships, and each node is intricately connected through multiple meta-paths. HAN uses a hierarchical attention mechanism to effectively reflect these structures and semantic features.
First, in node-level attention, the difference in importance between neighboring nodes is trained based on a given meta-path, and weights are assigned to nodes with higher relevance. Next, in semantic-level attention, a more important relationship for a task is trained among various meta-paths, and semantic embedding is integrated.

\section{Methodology}\label{sec:3.Methodology}

\begin{figure*}[t]
	\centering
    \includegraphics[width=\textwidth]{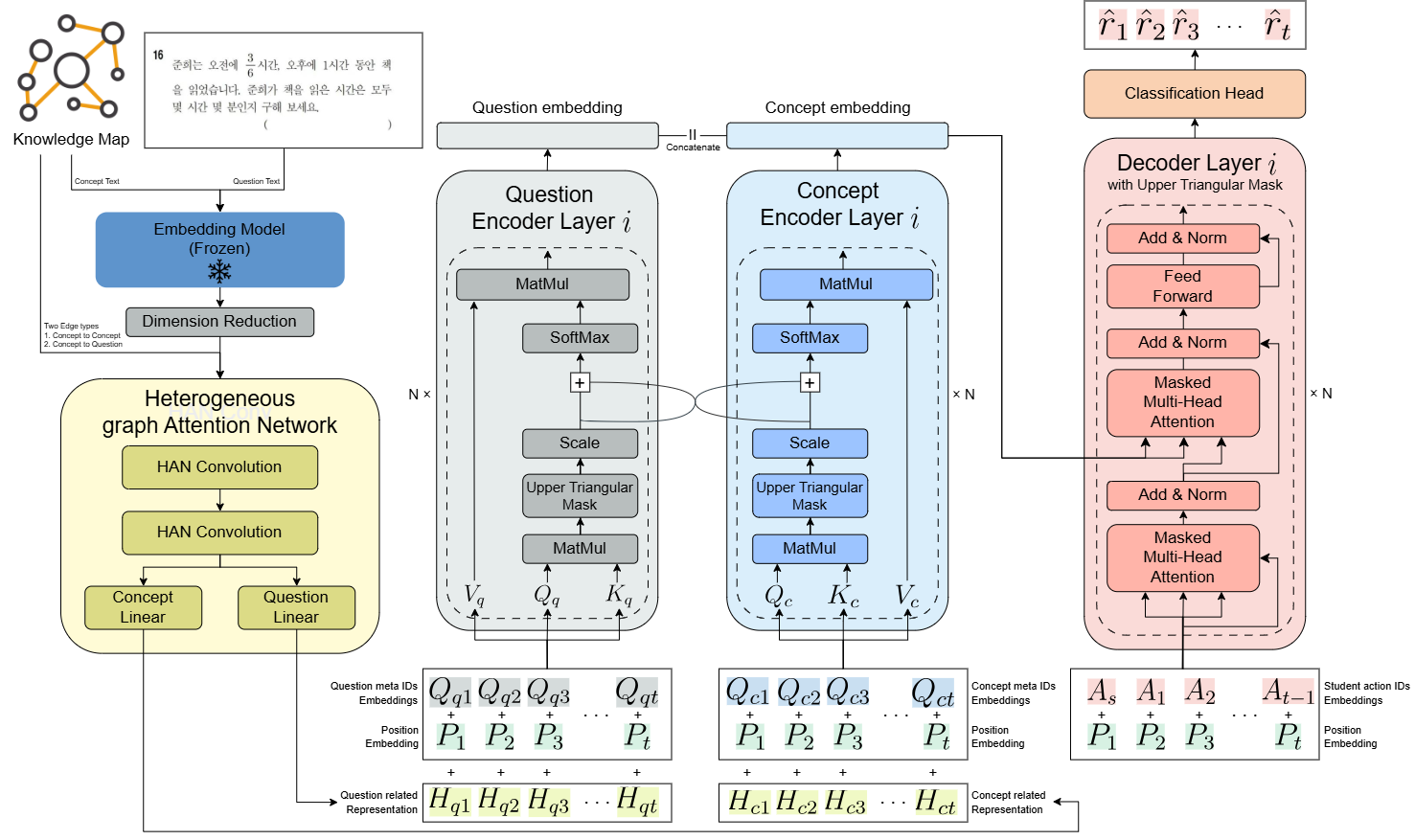}
	\caption{The architecture of Practical Integrated Cross-consistent Knowledge Tracing (PICKT). PICKT is a practical KT model for ITS services that estimates learners’ knowledge states by integrating diverse features while preserving their characteristics.}
	\label{FIG:1}
\end{figure*}

\subsection{Problem Definition}\label{sec:3.1.Problem Definition}

 In this study Practical Integrated Cross-consistent Knowledge Tracing (PICKT) model proposed expands the general settings of KT as shown in \protect\autoref{FIG:1}. PICKT utilizes various data elements in the sequence of interactions for each learner. Specifically, the input at each time point $t$ consists of meta feature ($Q_t$) related to the question, a fused texts and knowledge map ($H_t$) at that time point, and a learner response ($A_{t-1}$) to the previous point. Here, $Q_t$ is the feature of the question at time $t$, and $H_t$ is fused texts and knowledge map. Since the learner response ($A_t$) is not known at the time of response at time $t$, only the results up to time \( t{-}1 \) are considered. And at the beginning of the learner response sequence, it is initialized as a separate start token ($A_s$). In this way, PICKT infers future question responses by combining various data types in chronological order. By effectively reflecting a complex data structure, it provides a framework that can accurately predict the knowledge state of each learner.

\begin{equation}
  P(r_{t+1}=1 \mid I_1, I_2, ..., I_t, I_{t+1}),\; I_t=(Q_t, H_t, A_{t-1})
\label{eq:1}
\end{equation}

In other words, the learner’s historical data is given as {($Q_1$, $H_1$, $A_0$),…, ($Q_t$, $H_t$, $A_{t-1}$) as shown in Equation~\eqref{eq:1}. Based on this, the PICKT model aims to predict the probability of a correct response for a question at time \( t{+}1 \), utilizing the meta feature ($Q_{t+1}$), fused texts and knowledge map $(H_{t+1}$), and the accumulated learner responses up to time $t$ ($A_t$).

\subsection{PICKT Architecture}\label{sec:3.2.PICKT Architecture}

PICKT is a KT model that accurately estimates a learner's knowledge state by expanding and modifying the encoder-decoder structure based on transformer \citep{vaswani2017attention}. In particular, it exhibits excellent performance through a highly scalable structure that can effectively utilize a number of features. In this section, each component and structure of the PICKT model is described in detail.

\subsubsection{HAN Layer}\label{sec:3.2.1.HAN Layer}

We applied a Heterogeneous graph Attention Network (HAN) \citep{wang2019heterogeneous}, shown in the yellow part of \protect\autoref{FIG:1}. This network effectively incorporates the text of questions and concepts, along with a knowledge map that contains the relationships among concepts, and relationships between questions and concepts. The text of questions and concepts is represented using various pre-trained embedding models \citep{abimbola2026open}, including BERT \citep{reimers2019sentence}. To improve model efficiency, dimension reduction \citep{van2009dimensionality}, including PCA \citep{mackiewicz1993principal}, is applied after text embedding. In this paper, knowledge map features refer to the relational data contained within relationships among concepts, and relationships between questions and concepts.

In HAN Layer, we represent the knowledge map in a graph structure. Next, we combine text-based embeddings with the graph structure of a knowledge map. We adopted this approach because using question text alone has limitations in capturing question characteristics. From an educational perspective, questions are designed to target one or more related concepts \citep{junker2001cognitive}. The question–concept relationships and the sequential relationships among concepts reflect the achievement standards defined in the curriculum. Therefore, this study represents question characteristics by fusing the relational information from the knowledge map with the semantic information from the question texts. The HAN Layer is specifically designed to effectively train these two types of features.

Meanwhile, the PICKT does not train the text embedding model itself during the training process. Instead, it trains only a separate layer inside the HAN to effectively project the pre-trained embedding vector into the model's feature space. This is in line with the strategies used by the latest multi-modals such as LLaVA \citep{liu2023visual} and Flamingo \citep{alayrac2022flamingo} to effectively train large-scale data \citep{ji2025multimodal}. This study also adopted this training method to flexibly reflect texts feature in the model's substructure.

\subsubsection{Encoder Layer}\label{sec:3.2.2.Encoder Layer}

PICKT model was composed of two encoder layers by referring to the LiLT \citep{wang2022lilt} model so that data with different attributes can be effectively trained. Among the meta features related to the question, the feature ($Q_q$) of the question itself and the concept feature ($Q_c$) are calculated by another encoder layer and combined.

First, the Question Encoder Layer (the gray part of \protect\autoref{FIG:1}) trains the various characteristics of the question. It takes as input a feature ($Q_q$) that includes the question ID, question type, and difficulty level, along with the embedding corresponding to the question ID from the HAN Layer’s output, and effectively trains and incorporates the question’s texts and relationships of knowledge map ($H_q$). These various question related features were elaborately fused through a multi-layered transformation process including the self-attention mechanism. Consequently, the contextual relationships and complex interactions between the questions are effectively reflected.

Second, the Concept Encoder Layer (the blue part of \protect\autoref{FIG:1}) trains the various characteristics of the concept. It takes as input a feature ($Q_c$) that includes the concept ID, and the content, along with the embedding corresponding to the concept ID from the HAN Layer’s output, and effectively trains and incorporates the concept’s texts and relationships among concepts ($H_c$). These conceptual data are also integrated through a multi-layered transformation process including self-attention mechanisms. This process elaborately captures complex interactions and contextual links between concepts and reflects them deeply in the model.

In particular, in this study, when calculating the self-attention mechanism in the two encoder layers, unlike the existing LiLT structure, an upper triangular mask was applied. This ensures that model cannot access any information from time \( t{+}1 \) and beyond at the time $t$. This is to prevent information leakage that may occur when information at the future point is included in the process of estimating learners' knowledge state. This ensures that predictions are made based solely on information available up to the present, mirroring real-world learning environments. Therefore, in this study, the LiLT structure was modified and applied to meet this purpose.

The encoder of the PICKT model processes and combines question ($Q_q$) and concept ($Q_c$) features independently in each layer. And a masking technique was introduced in the self-attention mechanism within each encoder to ensure the causality of time series data. The combined operation of the overall model is defined as follows:

\begin{equation}
E = \sum_{m \in \{q, c\}} \mathrm{softmax}\left( \frac{Q_m K_m^\top}{\sqrt{d}} + Mask \right) V_m
\label{eq:2}
\end{equation}

In Equation~\eqref{eq:2}, $E$ denotes an integrated learner's representation, and \( m \in \{q, c\} \) denotes a question and concept encoders, respectively. In particular, the $Mask$ plays a role in improving the bidirectional attention structure of the existing LiLT model and blocking information of the future time point \( j \,(j > i) \) from being referenced at time point $i$. To achieve this, the model assigns $-\infty$ to future information while reflecting past and present information with weights of zero. Through this, it strictly prevents data leakage that could occur during the estimation of the learner's knowledge state. Finally, through the sum of the attention output values calculated by each encoder, an integrated expression that elaborately reflects the complex interaction between the question and the concept is derived.

\subsubsection{Decoder Layer}\label{sec:3.2.3.Decoder Layer}

Based on the integrated representation $E$ of questions and concepts derived from the encoder, the Decoder Layer combines the learner's response sequence ($A$) and estimates the final knowledge state.

The Decoder Layer of this model (shown in the red part of \protect\autoref{FIG:1}) performs a cross-attention mechanism. In this process, it uses the learner's response embedding sequence $R$ as the query, while utilizing the final output $E$ of the encoder as the key and value. In particular, like the encoder, an upper triangular $Mask$ was applied to the attention mechanism inside the decoder. This ensures that causal constraints are strictly maintained, preventing predictions at a specific point in time from depending on future learning data. This establishes a causal structure that forces the model to update the knowledge state based solely on previous response histories ($A$), question ($Q_q$, $H_q$) and concept ($Q_c$, $H_c$) features at every moment it predicts a learner's probability of a correct response. The decoder's output $Z_t$ is defined as follows:

\begin{equation}
Z_t = \mathrm{softmax}\left( \frac{R K_E^\top}{\sqrt{d}} + Mask \right) V_E
\label{eq:3}
\end{equation}

In Equation~\eqref{eq:3}, $R$ is a query matrix that embeds the learner's response feature, and $K_E$ and $V_E$ mean key and value matrices obtained by transforming the integrated output $E$ of the encoder.

\begin{equation}
\hat{y}_{t+1} = \sigma\left( f_2 \Big( \mathrm{Dropout} \big( \tanh \big( f_1(\mathrm{Dropout}(\mathbf{Z}_t)) \big) \big) \Big) \right)
\label{eq:4}
\end{equation}

As shown in Equation~\eqref{eq:4}, the final output $Z_t$ is processed through two linear neural network layers ($f_1$, $f_2$), an activation function, and a dropout layer to calculate $\hat{y}_{t+1}$, the final probability of the learner providing a correct response.

\subsubsection{Loss}\label{sec:3.2.4.Loss}

In this study, binary cross entropy loss \citep{ruby2020binary} was used to minimize the difference between the probability of correct answer to the question at the point $t$, which is the output of the model, and the actual correct answer result. Through this, the model trains to accurately predict the learner's probability distribution of correct answers. Specifically, the decoder's output $\hat{y}_t$ means the learner's probability of correct response about question $t$, which is then compared against the actual correctness \( y_t \in \{0, 1\} \). The total loss function $L$ is defined as follows:

\begin{equation}
\mathcal{L} = -\frac{1}{T} \sum_{t=1}^{T} \left[ y_t \log(\hat{y}_t) + (1 - y_t) \log(1 - \hat{y}_t) \right]
\label{eq:5}
\end{equation}

In Equation~\eqref{eq:5}, $T$ represents the total number of questions in the learning sequence. This loss function is optimized so that the model approaches the actual response distribution by handling the probability values in each case for correct and incorrect answers, and plays a key role in the accurate operation of the KT model.

\bigskip

Such a PICKT architecture provides a framework for precisely inferring individual learner's knowledge state by fusing texts and knowledge map features along with the difficulty feature of the question. In particular, this model has a high scalability to add and delete various features required in an educational service environment. This provides the structural advantage of being able to flexibly adjust the model according to actual service requirements.

This study aims to analyze which features are effective in securing the robustness of the model under the question cold start situation by utilizing the multifaceted data utilization capabilities of PICKT.

\section{Datasets}\label{sec:4.Datasets}

In this study, three datasets were used to evaluate the performance of multiple KT models and to examine their robustness under the question cold start situation. First, the MilkT (\url{https://www.milkt.co.kr}) dataset was used to assess the predictive performance under warm start, in which questions were observed during training. Second, the HME (\url{https://hme-ontest.chunjae.co.kr}) dataset was used to construct a strict question cold start situation, in which the questions used for evaluation were entirely held out from training. Third, the publicly available DBE-KT22 \citep{abdelrahman2022dbe} dataset was used as an external benchmark to examine whether the observed performance trends were consistent beyond the internally collected datasets. Summary statistics for each dataset are presented in \protect\autoref{TABLE:1}.

\begin{table*}[t] 
  \centering
  \caption{Statistics of the datasets.}
  \resizebox{\textwidth}{!}{%
    \begin{tabular}{cccccccc}
      \hline
      Dataset	& \# Questions	& \# Concepts & \# Learners & \# Interactions & Correct (\%) & Incorrect (\%) \\
      \hline
      MilkT & 16,026 & 100 & 75,848 & 16,773,700 & 11,842,647 (70.60\%) & 4,931,053 (29.40\%) \\
      HME & 100 & 53 & 578 & 14,450 & 9,711 (67.20\%) & 4,739 (32.80\%) \\
      DBE-KT22 & 212 & 98 & 1,264 & 161,645 & 123,631 (76.48\%) & 38,014 (23.51\%) \\
      \hline
    \end{tabular}
  }
  \label{TABLE:1}
\end{table*}

We have made our data publicly available to ensure the reproducibility of our research
(\url{https://huggingface.co/datasets/wonbeeny/milkt-hme-v1}).

\subsection{MilkT}\label{sec:4.1.Online}

The MilkT dataset was constructed from interaction logs between learners and questions collected from an online learning platform operated by our company (\url{https://www.chunjae.co.kr}), a major domestic edtech company. The platform is designed to provide practice questions after concept lectures, allowing learners to internalize the concepts they have just learned. Consistent with this service design, the dataset includes a relatively large number questions of easy to moderate difficulty level intended for foundational learning. The dataset covers the first semester of 2025 and includes interaction logs generated by students in grades 3–6 within the elementary mathematics curriculum.

Question metadata consisted of grade, concept, the type, the activity, the content, text, and difficulty. However, the source data contained missing values in text and in some metadata fields. In particular, text was available only as image format and therefore required Optical Character Recognition (OCR) \citep{islam2017survey} for text extraction. To address this limitation, question text was automatically extracted from the images using a Vision Language Model (VLM) \citep{zhang2024vision}. A content-concept relation table was then constructed using detailed definitions of concept and content categories as inputs to the VLM. Each concept was assigned to a single content category based on the official curriculum guidelines. The knowledge map contained a total of 100 concepts, each uniquely associated with a specific combination of grade and the content. For each question, grade and the content were first used together with the predefined content–concept relation table to narrow the set of candidate concepts. The VLM was then provided with both the question image and the previously extracted text and instructed to select the single concept most essential for solving the question from the narrowed candidate set. To assess the reliability of the concept assigning procedure, 200 samples were selected from each grade and reviewed by domain experts. The content–concept relation table was also verified by the domain human experts. The Qwen/Qwen3-VL-30B-A3B-Instruct \citep{bai2025qwen3} model was used for text extraction and VLM-assisted metadata completion, with specific examples provided in ~\ref{app:b.Multi-step Data Generation with VLM Prompts}. By contrast, the activity, the type, and difficulty were retained as originally annotated by experts without additional correction, because these fields contained no missing values.

Detailed definitions of each feature are as follows:

\begin{itemize}
  \item concept: The knowledge component required to solve a specific question. Each concept was defined by domain experts with reference to official curriculum and textbooks of Korea \citep{moe2022mathcurriculum}.
  \item grade : The curriculum grade level associated with the question.
  \item question text: Texts of the question, including the problem and the explanation. For easy questions, an explanation may not be provided.
  \item activity: The four categories of cognitive demand required to solve the question. The categories were based on the criteria used by the Korea Institute for Curriculum and Evaluation \citep{kice2026csatstudyguide}.
  \item content: The four categories of the concepts, as defined in the official curriculum and textbooks of Korea. Each concept was assigned to one of these content areas.
  \item type: The response format required by the question, classified as multiple-choice, short-answer, or etc.
  \item difficulty: Difficulty level of the question annotated by human experts. It has five levels.
  \item knowledge map: An expert-designed graph based on the official curriculum of Korea. It links each question to its assigned concept and represents prerequisite relationships among concepts.
\end{itemize}

\subsection{HME}\label{sec:4.2.Offline}

The HME that is offline math competition dataset was collected independently from the online platform data (MilkT). Our company holds an annual HME, and the 2025 competition was held on June 14. The competition questions were developed exclusively for the competition and were not used on the online platform. Therefore, from the perspective of a model trained only on MilkT data, the competition questions constitute previously unseen questions. These questions share the same metadata schema as the online platform questions. In addition, a subset of learners had both online learning histories and competition participation records. Leveraging this overlap, this study first trained the model using only the learning history from the MilkT data and restricted the analysis to the overlapping learners. Subsequently, it simulated the question cold start scenario by appending HME questions to their prior MilkT activity.

\subsection{DBE-KT22}\label{sec:4.3.DBE-KT22}

The DBE-KT22 \citep{abdelrahman2022dbe} data is a KT benchmark dataset collected from students enrolled in an introductory relational database course at the Australian National University (ANU) between 2018 and 2021. This dataset provides various question metadata such as question text and expert-assigned difficulty information. In particular, the dataset includes text of each question problem, enabling the training of natural language embeddings. Furthermore, it provides a knowledge map that explicitly defines the relationship structure between questions and concepts as well as the relationships among concepts. This relationship structure makes it suitable for the application of graph neural network based models.

\section{Experimental Setup}\label{sec:5.Experimental Setup}

\subsection{Baseline Models}\label{sec:5.1.Baseline Models}

DKT (Deep Knowledge Tracing) \citep{piech2015deep}. It uses concepts and responses as input data. As an RNN-based \citep{schmidt2019recurrent} model, it represents a learner’s knowledge state as hidden state vectors and updates the probability of the next correct response at each time step.

DKVMN (Dynamic Key-Value Memory Networks) \citep{zhang2017dynamic}. It adopts key and value matrices. The key matrix stores concept-specific weights for a given question, while the value matrix tracks a learner’s evolving achievement level for each concept, updating it based on responses.

SAKT (Self-Attentive Knowledge Tracing) \citep{pandey2019self}. It applies the self-attention mechanism of transformers to estimate the probability of a correct response by calculating the relative importance of previously solved questions.

GKT (Graph-based Knowledge Tracing) \citep{nakagawa2019graph}. It models concepts and their relationships as a graph. It updates both the target concept and proximate concepts, enabling precise learner achievement estimation and probability calculation for related questions.

AKT (Attentive Knowledge Tracing) \citep{ghosh2020context}. By utilizing Monotonic Attention, AKT adjusts weights over time to modulate the amount of information, thereby reflecting the characteristics of time-series learning and the decay of memory. Through Rasch \citep{rasch1993probabilistic, von2016rasch} model-based regularization, AKT trains the model so that differences in question difficulty within the same concept are reflected in the question/concept embeddings, allowing it to naturally incorporate question difficulty information into the model representation.

SAINT (Separated Self-Attentive Neural Knowledge Tracing) \citep{choi2020towards}. It adopts a Transformer encoder-decoder architecture to process the question sequence and the learner’s response sequence separately. The encoder uses information related to the question, while the decoder uses whether the learner answered the question correctly or incorrectly.

DTransformer (Diagnostic Transformer) \citep{yin2023tracing}. It diagnoses achievement at the question level and tracks overall knowledge states, ensuring stable learner knowledge representation via contrastive learning.

\begin{figure*}[t]
	\centering
    \includegraphics[width=1.2\columnwidth]{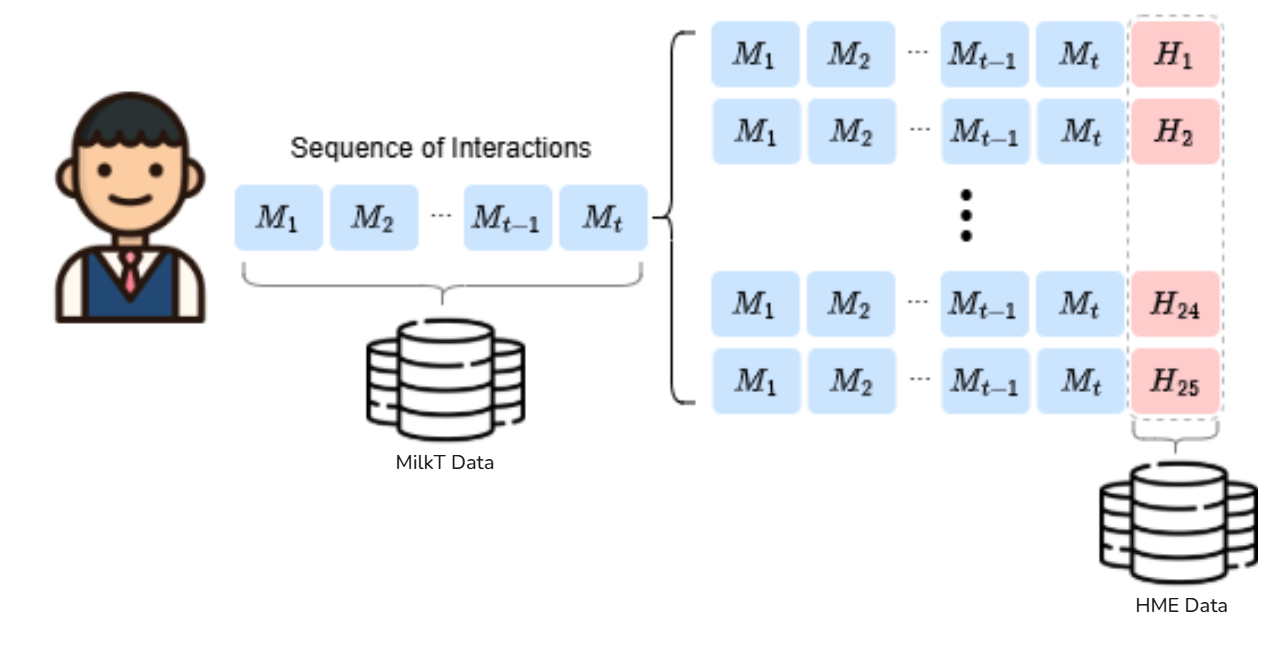}
	\caption{The preprocessing of the interaction sequences for the cold start simulation.}
	\label{FIG:2}
\end{figure*}

\subsection{Model Training and Evaluation}\label{sec:5.2.Model Training and Evaluation}

The data was divided into training and validation datasets, at an 8:2 ratio for both the MilkT and DBE-KT22 datasets. The optimization and hyperparameter settings used for model training are as follows. For the PICKT model, we used the ADAM optimizer \citep{kingma2014adam} (learning rate 3e-4, epsilon 1e-12), the GELU activation \citep{hendrycks2016gaussian} function, dropout of 0.1, and weight initialization based on a normal distribution with a mean of 0 and a standard deviation of 0.02. For the MilkT dataset, the maximum input sequence length, batch size, and number of epochs were set to [100, 128, 5], while for the DBE-KT22 dataset, they were set to [32, 32, 20]. For baseline comparisons, the selected KT models were configured with the same sequence length, batch size, and epoch settings as PICKT, while the remaining hyperparameters followed the specifications in the original papers. Training and evaluation for all models were performed on an NVIDIA A100-SXM4-40GB GPU.

Model performance was evaluated using four metrics. The dataset used in this study exhibits class imbalance \citep{he2009learning, yadav2020handling}, with the number of correct answers exceeding that of incorrect answers. Because of this imbalance, overall accuracy alone may not provide a fair evaluation of the model’s predictive performance \citep{saito2015precision}. Therefore, we separately measured the models’ predictive performance for learners’ actual correct and incorrect responses. This allowed us to assess whether the models could recognize patterns of both correct and incorrect responses \citep{bradley1997use, sokolova2009systematic}. To classify answers as correct or incorrect, a threshold of 0.5 was applied. Predictions with a probability of 0.5 or higher were classified as correct, while those below 0.5 were classified as incorrect. As shown below, various metrics \citep{fawcett2006introduction} were used to evaluate the predictive performance of the KT models in a more detailed and reliable manner.

\begin{itemize}
  \item Incorrect Accuracy (ACC Incorrect): The proportion of learners’ observed incorrect responses that the model correctly predicted, evaluating the model’s ability to recognize the pattern of incorrect responses. This is equivalent to Specificity (True Negative Rate).
  \item Correct Accuracy (ACC Correct): The proportion of learner’s actual correct responses that the model correctly predicted, evaluating the model’s ability to recognize the pattern of correct responses. This is equivalent to Sensitivity (True Positive Rate).
  \item Accuracy (ACC): The proportion of interactions that were correctly classified by the model. This metric provides an overall evaluation of the model’s predictive performance across the entire dataset.
  \item Area Under the roc Curve (AUC) \citep{bradley1997use}: This metric indicates how well the model’s predictions distinguish between actual correct and incorrect responses, providing a comprehensive evaluation of the model’s overall discriminatory ability.
\end{itemize}

We have made our code publicly available to ensure the reproducibility of our research
(\url{https://github.com/wonbeeny/PICKT/tree/main}).

\subsection{Cold Start Experiment}\label{sec:5.3.Cold Start Experiment}

As described in Section~\ref{sec:4.2.Offline}, this study simulated the question cold start situation by combining the MilkT and HME interaction sequences of the same learner in chronological order. During the training phase of the KT model, we strictly ensured that the model had no prior exposure to the questions contained in the HME data by using only MilkT questions.

The specific procedure for generating the question cold start scenario dataset is illustrated in \protect\autoref{FIG:2}. First, the learner’s MilkT sequences of interactions ($M_1$ to $M_t$) was replicated 25 times to match the number of HME questions (25). Then, one HME question ($H_1$ to $H_{25}$) was appended to the end of each copy, creating 25 independent sequences of length \( t{+}1 \). The HME sequences of interactions was generated from a math competition held on June 14, 2025, at 11:30 AM, and the MilkT sequences of interactions was limited to data from May 31 to June 13, 2025. This is based on prior research indicating that a learner’s knowledge state remains stable without significant change over a period of approximately two weeks \citep{valderama2021learning}. Subsequently, only the model’s predictions for the question at time \( t{+}1 \)—the final point in the sequence—were used for performance evaluation.

This design replicates a scenario where only the last question is excluded from the training data. This enables a more realistic evaluation of the model’s performance under the question cold start situation on real-world educational platforms. To ensure a rigorous evaluation, the model was trained without using any MilkT or HME interaction sequences from learners included in the cold start scenario set.

\section{Results}\label{sec:6.Results}

This section presents the results of comprehensive experiments designed to demonstrate the robustness of the model when new questions are introduced. The purpose of these experiments goes beyond simply comparing predictive performance. It is to empirically identify which features the KT model should utilize to address the question cold start problem. Through this, we aim to verify that the KT model can provide reliable learning assessment results under real-world educational services in which new questions are routinely introduced. The specific experimental analysis and discussion are conducted around the following four research questions:

\begin{itemize}
  \item RQ1: Verify the predictive performance of PICKT by comparing it with baseline KT models.
  \item RQ2: Verify PICKT’s ability to stable robust under the question cold start scenario.
  \item RQ3: Identify the key features that contribute to the performance improvement of the PICKT model under the question cold start scenario through ablation experiments.
  \item RQ4: Experimentally demonstrate how the key features identified in RQ3 affect the robustness of the PICKT model under the question cold start scenario.
\end{itemize}

\subsection{Prediction Performance (RQ1)}\label{sec:6.1.Prediction Performance (RQ1)}

In this study, we compared the overall predictive performance of PICKT and baseline KT models using MilkT and DBE-KT22 datasets. Our aim is to validate the effectiveness of PICKT in tracing learners’ knowledge states under warm start conditions. The performance of each model across the evaluation metrics is presented in \protect\autoref{TABLE:2}.

\begin{table*}[t]
\centering
\caption{Predictive performance under warm start between PICKT and baseline KT models on the MilkT and DBE-KT22 datasets.}
\label{TABLE:2}
\resizebox{\textwidth}{!}{%
\begin{tabular}{l cccc cccc}
\toprule
& \multicolumn{4}{c}{MilkT} 
& \multicolumn{4}{c}{DBE-KT22} \\
\cmidrule(lr){2-5}\cmidrule(lr){6-9}
Model & ACC Incorrect & ACC Correct & ACC & AUC & ACC Incorrect & ACC Correct & ACC & AUC \\
\midrule
DKT & 0.4865±0.0024 & 0.9158±0.0005 & 0.7891±0.0007 & 0.8194±0.0008 & 0.2686±0.0097 & 0.9474±0.0021 & 0.7861±0.0050 & 0.7652±0.00474 \\
DKVMN & 0.4150±0.0027 & \textbf{0.9216±0.0004} & 0.7725±0.0007 & 0.7866±0.0008 & 0.1214±0.0037 & 0.9717±0.0017 & 0.7696±0.0055 & 0.7078±0.0056 \\
SAKT & 0.4594±0.0037 & 0.9106±0.0010 & 0.7778±0.0005 & 0.7985±0.0009 & 0.2487±0.0074 & 0.9499±0.0024 & 0.7832±0.0048 & 0.7633±0.0045 \\
GKT & 0.4307±0.0127 & 0.9177±0.0056 & 0.7744±0.0007 & 0.7910±0.0010 & 0.0555±0.0098 & \textbf{0.9888±0.0030} & 0.7670±0.0049 & 0.6426±0.0072 \\
AKT & 0.5574±0.0033 & 0.9114±0.0009 & 0.8073±0.0003 & 0.8513±0.0004 & 0.3364±0.0098 & 0.9335±0.0025 & 0.7916±0.0048 & 0.7919±0.0041 \\
SAINT & 0.5490±0.0032 & 0.9148±0.0009 & 0.8071±0.0002 & 0.8509±0.0005 & 0.3305±0.0112 & 0.9330±0.0059 & 0.7899±0.0049 & 0.7881±0.0032 \\
DTransformer & 0.5375±0.0048 & 0.9146±0.0014 & 0.8036±0.0003 & 0.8457±0.0007 & 0.2956±0.0096 & 0.9421±0.0019 & 0.7884±0.0053 & 0.7825±0.0033 \\
PICKT & \textbf{0.5675±0.0015} & 0.9138±0.0005 & \textbf{0.8119±0.0002} & \textbf{0.8579±0.0004} & \textbf{0.3496±0.0094} & 0.9331±0.0017 & \textbf{0.7945±0.0043} & \textbf{0.7979±0.0035} \\
\bottomrule
\end{tabular}
}
\end{table*}

As shown in \protect\autoref{TABLE:2}, PICKT achieved the highest scores among all benchmark models in both ACC (0.8119) and AUC (0.8579) on the MilkT dataset, indicating that it traces learners’ knowledge states more accurately than the baseline models. In particular, PICKT showed a clear advantage in predicting incorrect responses, as reflected in ACC Incorrect. PICKT recorded an ACC Incorrect of approximately 0.5675, outperforming AKT (0.5574), the strongest baseline model for this metric, by 1.01\% and exceeding several other baseline models by more than 15\%. These results suggest that PICKT is particularly effective in identifying learners’ incorrect response patterns, which may contribute to a more precise diagnosis of learners’ weaknesses.

PICKT’s superior performance was also observed in experiments using the DBE-KT22 dataset. On DBE-KT22, PICKT achieved the highest scores in both ACC (0.7945) and AUC (0.7979), suggesting that the model maintains strong predictive performance across different datasets. In particular, whereas the baseline models showed relatively low ACC Incorrect values, ranging from 0.05 to 0.33, PICKT recorded the highest ACC Incorrect of 0.3496, indicating its stronger ability to predict actual incorrect responses. This value exceeded those of AKT (0.3364) and SAINT (0.3305), the strongest baseline models for this metric, by 1.32\% and 1.91\%, respectively. These results suggest that PICKT maintains robust performance in identifying incorrect responses even when evaluated on a different dataset, which may contribute to a more reliable diagnosis of learners’ weaknesses.

\subsection{Cold Start (RQ2)}\label{sec:6.2.Cold Start (RQ2)}

We conducted a simulated the question cold start scenario to evaluate how stably each model maintains its predictive performance when new questions unseen during training are introduced. Following the design described in Section~\ref{sec:5.3.Cold Start Experiment}, the experiment was conducted by appending one unseen HME question to the end of the MilkT interaction sequences. The model then predicted whether the learner would answer the appended question correctly or incorrectly. The performance of each model under the question cold start scenario is presented in \protect\autoref{TABLE:3}.

\begin{table}[h!]
\centering
\renewcommand{\arraystretch}{1.2}  
\caption{Comparison of predictive performance between PICKT and baseline KT models under the question cold start scenario.}
\label{TABLE:3}
\resizebox{\linewidth}{!}{%
\begin{tabular}{ccccc} \hline
Model & ACC Incorrect & ACC Correct & ACC & AUC \\ \hline
DKT & 0.2802 & 0.8273 & 0.6479 & 0.6330 \\
DKVMN & 0.3186 & 0.8047 & 0.6453 & 0.6213 \\
SAKT & 0.3060 & 0.8084 & 0.6436 & 0.6219 \\
GKT & 0.2211 & 0.8767 & 0.6617 & 0.6284 \\
AKT & 0.2739 & 0.8498 & 0.6609 & 0.6413 \\
SAINT & 0.3053 & 0.8115 & 0.6455 & 0.6288 \\
DTransformer & 0.3446 & 0.7817 & 0.6383 & 0.6061 \\
PICKT & \textbf{0.4117} & \textbf{0.9521} & \textbf{0.7749} & \textbf{0.8350} \\
\hline
\end{tabular}%
}
\end{table}

As shown in \protect\autoref{TABLE:3}, the experimental results indicate that baseline KT models exhibited a sharp decline in performance when faced with new questions that had been excluded from the training data. In contrast, PICKT maintained higher predictive performance than the baseline models across all evaluation metrics, including ACC (0.7749) and AUC (0.8350). These results suggest that PICKT mitigates the limitations of methods that rely primarily on question specific identifiers (question IDs) by effectively utilizing unique information inherent to each question. In particular, whereas the baseline models showed substantial decreases in ACC Correct and ACC Incorrect under the question cold start scenario, PICKT maintained stable performance, recording 0.9521 for ACC Correct and 0.4117 for ACC Incorrect.

Consequently, PICKT not only achieved strong performance under warm start, as examined in RQ1, but also showed high reliability under the question cold start scenario, as examined in RQ2. Based on this experimental evidence, this study uses PICKT as the target model for ablation study to identify which features make critical contributions to its robustness under the question cold start scenario.

\subsection{Ablation Study (RQ3)}\label{sec:6.3.Ablation Study (RQ3)}

This section examines the key features that contribute to PICKT’s predictive performance. Motivated by the prior studies discussed in Section~\ref{sec:2.2.Cold Start}, this study focuses on difficulty, text, and knowledge map features. The knowledge map encodes both relationships among concepts and relationships between questions and concepts. In PICKT, text and relational information from the knowledge map are fused through the HAN layer as described in Section~\ref{sec:3.2.1.HAN Layer} Accordingly, we denote the resulting fused representation as the HAN.

The experiments consist of ablation studies under two scenarios. First, using the same evaluation method as in RQ1, we measured model performance on questions included in the MilkT dataset, rather than on newly introduced questions.

\begin{table}[h!]
  \centering
  \caption{Comparison of PICKT’s predictive performance according to the selective exclusion of key features on the MilkT dataset.}
  \label{TABLE:4}
  \footnotesize  
  \renewcommand{\arraystretch}{1.2}  
  \begin{tabular}{lcccc} 
    \hline
    Model & ACC Incorrect & ACC Correct & ACC & AUC \\ 
    \hline
    w/o both & 0.5547 & \textbf{0.9182} & 0.8111 & 0.8569 \\
    w/o HAN & 0.5681 & 0.9133 & 0.8116 & 0.8576 \\
    w/o Diff & \textbf{0.5707} & 0.9121 & 0.8116 & 0.8576 \\
    PICKT & 0.5692 & 0.9129 & 0.8116 & \textbf{0.8577} \\
    \hline
  \end{tabular}
  
  \vspace{2mm}
  \raggedright
  \scriptsize  
  \textbf{Note.} w/o both: PICKT without HAN layer's features and difficulty feature; w/o HAN: PICKT without HAN layer's features; w/o Diff: PICKT without difficulty feature.
\end{table}

As shown in \protect\autoref{TABLE:4}, when evaluated on questions seen during training, the ablated variants showed no substantial performance degradation compared with the full PICKT. w/o Diff, w/o HAN, and w/o Both achieved comparable performance, despite removing the difficulty feature, the HAN, or both, respectively.

To examine whether this performance is maintained under the question cold start scenario, the central focus of this study, we conducted an ablation study under the same cold start scenario used for RQ2. The changes in model performance resulting from the exclusion of key features under the scenario are presented in \protect\autoref{TABLE:5}.

\begin{table}[h!]
\centering
\renewcommand{\arraystretch}{1.2}  
\caption{Comparison of PICKT’s predictive performance according to the exclusion of key features on unseen HME questions under the question cold start scenario.}
\label{TABLE:5}
\resizebox{\linewidth}{!}{%
\begin{tabular}{ccccc} \hline
Model & ACC Incorrect & ACC Correct & ACC & AUC \\ \hline
w/o both & 0.1277 & 0.9548 & 0.6835 & 0.6799 \\
w/o HAN & \textbf{0.4391} & 0.9432 & \textbf{0.7779} & 0.8242 \\
w/o Diff & 0.1452 & \textbf{0.9561} & 0.6902 & 0.7151 \\
PICKT & 0.4060 & 0.9517 & 0.7727 & \textbf{0.8375} \\
\hline
\end{tabular}%
}
\end{table}

As shown in \protect\autoref{TABLE:5}, the ablation results show that removing key features led to clear performance degradation. w/o Both showed the largest decline, with AUC decreasing to 0.6799, the lowest among the variants. Performance also degraded when either component was removed individually. w/o HAN recorded an AUC of 0.8242, whereas w/o Diff recorded an AUC of 0.7151. The higher performance of the single-ablation variants compared with w/o Both suggests that both the difficulty feature and the HAN features contribute meaningfully to maintaining PICKT’s predictive performance and robustness under the question cold start scenario.

The following section further examines the educational scenarios in which the influence of each feature becomes more pronounced. Because the distribution of difficulty level of questions is closely related to the characteristics of educational services \citep{lord1952relation}, an analysis by difficulty level interval is necessary to determine which feature should be prioritized under different service characteristics. Accordingly, the subsequent analysis provides empirical evidence for selecting the most relevant feature according to the characteristics of each scenario.

\subsection{Difficulty and HAN Effects under the Question Cold Start Scenario (RQ4)}\label{sec:6.4.Difficulty and HAN Effects under Cold Start (RQ4)}

We analyze how the difficulty and the HAN identified in RQ3 affect PICKT’s robustness under the question cold start scenario. To this end, we compared the model’s performance across different difficulty levels. Because the HME data included only four difficulty levels (easy, medium, hard, very hard), the very easy difficulty level was excluded from the analysis.

\begin{table}[h!]
  \centering
  \caption{Comparison of PICKT’s predictive performance by difficulty levels after excluding key features.}
  \label{TABLE:6}
  \resizebox{\columnwidth}{!}{%
    \begin{tabular}{l l cccc}
      \toprule
      Difficulty & Model & ACC Incorrect & ACC Correct & ACC & AUC \\
      \midrule
      \multirow{2}{*}{Easy}
       & w/o HAN & 0.0421 & \textbf{0.9800} & \textbf{0.9209} & 0.6396 \\
       & w/o Diff & \textbf{0.0955} & 0.9671 & 0.9122 & \textbf{0.6651} \\
      \hdashline
      \multirow{2}{*}{Medium}
       & w/o HAN & 0.1437 & 0.9268 & 0.7244 & 0.5989 \\
       & w/o Diff & \textbf{0.1476} & \textbf{0.9428} & \textbf{0.7372} & \textbf{0.6034} \\
      \hdashline
      \multirow{2}{*}{Hard}
       & w/o HAN & \textbf{0.3070} & 0.8018 & \textbf{0.4684} & 0.6068 \\
       & w/o Diff & 0.1421 & \textbf{0.9266} & 0.3980 & \textbf{0.6316} \\
      \hdashline
      \multirow{2}{*}{Very Hard}
       & w/o HAN & \textbf{0.8579} & 0.0889 & \textbf{0.8379} & \textbf{0.4761} \\
       & w/o Diff & 0.1788 & \textbf{0.7111} & 0.1926 & 0.3931 \\
      \bottomrule
    \end{tabular}
  }
\end{table}

As shown in \protect\autoref{TABLE:6}, the difficulty feature played an important role in maintaining PICKT’s predictive performance for questions at the very hard difficulty level. Specifically, w/o HAN showed high ACC Incorrect but low ACC Correct for questions with the very hard difficulty level. This is because the correct response rate for that level was very low at 2.6\%, leading the model to predict mostly incorrect response. Conversely, for easy difficulty level questions, ACC Incorrect was low and ACC Correct was high, as the correct response rate in that level was high at 93.7\%, leading the model to predict mostly correct responses. In other words, while the difficulty feature provides strong robustness when the difficulty level is at the very hard, it also tends to cause the model’s predictions to be heavily biased toward the correct response rate.

In contrast, w/o Diff showed higher AUC than w/o HAN in the easy, medium, and hard difficulty levels. This indicates that the HAN enabled the model to utilize fused texts and knowledge map features to capture the characteristics of the questions themselves. In the training dataset (MilkT), questions were distributed mainly across medium difficulty level (45.37\%), easy difficulty level (40.17\%), and hard difficulty level (14.20\%), making it likely that questions structurally or semantically similar to the newly introduced questions were included in the training data. As a result, even for the question cold start scenario, the model appears to have reflected the structural and textual-related characteristics of the questions. However, because the training data contained only 28(0.17\%) very hard difficulty level questions, w/o Diff showed degraded performance in this level, likely due to insufficient exposure to similar questions during training.

In summary, the difficulty feature provides a strong predictive signal for very hard difficulty level questions, but prediction bias may arise due to the correct response rate of extremely difficult questions. In addition, the HAN captures question-specific characteristics through textual and relational information fused in the HAN layer, thereby supporting more robust performance under the question cold start scenario when questions similar to those in the training data are available.

\section{Conclusion}\label{sec:7.Conclusion}

In this study, we empirically identified the key features that secure the robustness of PICKT under the question cold start scenario, where new questions are continuously introduced. The experimental results confirmed that the difficulty and HAN (Section~\ref{sec:3.2.1.HAN Layer}) features play complementary roles and maintain the model’s predictive performance. Their relative contributions vary across difficulty of questions. These findings offer empirical implications for designing KT models tailored to the characteristics of real-world ITS services. 

First, in services such as math competition preparation involving a substantial proportion of questions with very hard difficulty level, it is necessary to prioritize the difficulty feature. This is because when the difficulty of a question is extremely hard, it has a strong correlation with the question’s correct response rate \citep{quaigrain2017using, rudolph2019best}. Therefore, it is necessary to consider the potential bias arising from the question’s correct response rate. \\
Second, in services focused on foundational to intermediate learning, it is necessary to prioritize the fused texts and knowledge map features. This is because many questions at these difficulty levels are likely to share similar textual or relational characteristics \citep{byrd2022predicting, benedetto2020r2de, alkhuzaey2024text}. However, it should also be considered whether there are actually sufficient similar questions.

In conclusion, when designing real-world ITS services, it is essential to prioritize the selection and preparation of appropriate features according to the educational objectives of the service. By applying the feature selection principles proposed in this study, stable diagnostic performance can be achieved even under the question cold start situations, thereby enabling the provision of more sophisticated personalized learning paths for learners.

\section{Limitation}\label{sec:8.Limitation}

While this study achieved meaningful results in identifying features that support the robustness of KT models under the question cold start situations, it also has several limitations that should be addressed in future research to enable more immediate adaptation to diverse scenarios in real-world educational settings.

First, preparing the metadata required to enhance predictive performance involves considerable human effort. The difficulty, texts, and relational information of knowledge map utilized in this study often require detailed manual annotation \citep{moon2024using} and validation by domain experts. To streamline this process, we are conducting follow-up research on the following automation techniques.

\begin{itemize}
  \item Optimization of difficulty estimation: In the cold start scenario examined in this study, newly introduced questions have no interaction history, making it difficult to estimate their difficulty using IRT-based \citep{wolins1982probabilistic} methods. To address this limitation, we are investigating a method that estimates the difficulty of new questions by measuring their similarity to existing questions and utilizing the difficulty feature of highly similar questions \citep{sarwar2001item}.
  \item Automatic knowledge structure generation: The knowledge map used in this study was manually constructed by human experts who defined relationships among concepts. To automate this process \citep{zhong2023comprehensive}, we aim to build a more extensive and sophisticated knowledge structure by automatically extracting concept nodes and edges \citep{chen2018knowedu} through an ontology-based knowledge graph \citep{maedche2001ontology, chen2018automatic, asim2018survey}.
\end{itemize}

Second, the robustness of the model observed in this study needs to be further validated across diverse educational domains \citep{jung2024clst}. The experiments were conducted with mathematics data from elementary school students. Although the results demonstrate the validity of the proposed approach under these specific conditions, future research should expand the evaluation scope to improve the model’s generalizability.

\begin{itemize}
  \item Diversification of subjects and grade levels: Future studies will expand the experimental scope to examine how differences in question difficulty across elementary, middle school, high school, and university levels, as well as subject-specific characteristics beyond mathematics, affect the model’s predictive performance.
\end{itemize}

In conclusion, future research will focus on developing a personalized ITS that maximizes service scalability while substantially reducing reliance on human labor by integrating an LLM-based metadata auto-extraction system with an ontology-based knowledge structure.

\section{Statements on open data and ethics}\label{sec:9.Statements on open data and ethics}

The DBE-KT22 dataset used in this study is available for public access through the Australian Data Archive platform (\url{https://dataverse.ada.edu.au/dataset.xhtml?persistentId=doi:10.26193/6DZWOH}). Additional information on the dataset is provided in the related article. Our company’s datasets analyzed in this study is publicly available on Hugging Face Hub (\url{https://huggingface.co/datasets/wonbeeny/milkt-hme-v1}). The data and the codes that support the findings of the study are stored in GitHub (\url{https://github.com/wonbeeny/PICKT/tree/main}). All analyses and procedures were performed in compliance with the relevant laws and institutional guidelines. Strict measures are taken to protect the privacy of involved students.

\appendix
\renewcommand{\thesection}{Appendix \Alph{section}}
\section{Overview of Missing Metadata Reconstruction}\label{app:a.Overview of Missing Metadata Reconstruction}

\begin{table*}[t]
  \centering
  \renewcommand{\thetable}{A.\arabic{table}}
  \setcounter{table}{0}
  \caption{Overview of missing value processing, showing the VLM inputs, reconstructed outputs, and validation strategy for each feature.}
  \label{TABLE:A.1}
  \footnotesize
  \begin{tabular}{l p{5.3cm} p{2.2cm} p{4cm}} 
    \toprule
    Feature & Input to VLM & Output in Data & Validation \\ 
    \midrule
    Question text & Each image of Problem and Explanation & OCR-style text & Expert review on sampled cases \\
    Content & Question (image \& text) + Content list & Assigned content & Expert review on sampled cases \\
    Content-Concept relations & Content list + Concept & Relation table & Expert review on full cases \\
    Concept & Question (image \& text) + Concept list & Assigned concept & Expert review on sampled cases \\
    \bottomrule
  \end{tabular}
\end{table*}

\protect\autoref{TABLE:A.1} summarizes the overall procedure for handling missing values that occurred in the data generation process. The brief processing process is as follows:

\begin{itemize}
  \item Question text: Extracted using the question image as input.
  \item Content: Established by utilizing both the question image and the extracted text.
  \item Content-concept relationships: Established a relationship table for all concepts by predicting which concept corresponds to each content.
  \item Concept: Established by filtering the list of concepts that can be selected based on the grade and content of the question, and then utilizing all information.
  \item Others: Use the original value with other metadata written by our company human experts without additional processing.
\end{itemize}

For data construction, the Qwen/Qwen3-VL-30B-A3B-Instruct \citep{bai2025qwen3} model was used as mentioned in Section~\ref{sec:4.1.Online}. Before the final reflection, reliability was secured through an human experts verification procedure.

\section{Multi-step Data Generation with VLM Prompts}\label{app:b.Multi-step Data Generation with VLM Prompts}

\begin{figure*}[!t]
  \centering
  \renewcommand{\thefigure}{B.\arabic{figure}}
  \setcounter{figure}{0}
  \includegraphics[width=\textwidth]{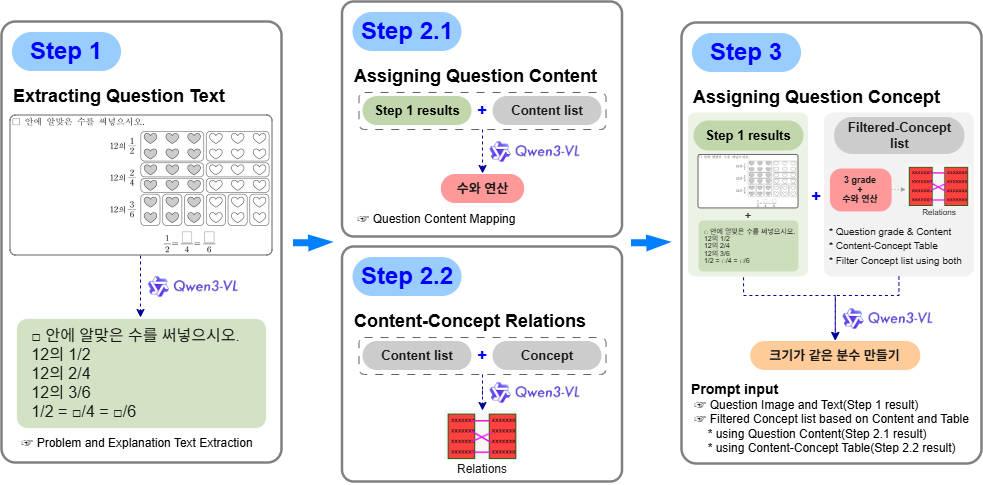}
  \caption{Step-by-step Qwen3-VL-based procedure for extracting question text and assigning predefined content and concept labels using question images and content-concept relations.}
  \label{FIG:B.1}
\end{figure*}

For the treatment of missing values of question text, the content, and concept, a three-steps procedure was performed as shown in \protect\autoref{FIG:B.1}.

In Step 1, text was extracted from the problem and explanation of the question saved as images, and the prompts (\hyperlink{step1}{Step 1}) used are provided at the bottom.

Step 2 consists of two steps. In Step 2.1, one of the predefined contents was assigned using the question image and the problem and explanation texts extracted in Step 1. As mentioned in the Section~\ref{sec:4.1.Online}, the content is the same as the content system of the official curriculum of Korea. Therefore, the description specified in the curriculum was provided together so that the VLM could make a more specific judgment. The prompt (\hyperlink{step2.1}{Step 2.1}) to assign the content to the question is provided at the bottom.

In Step 2.2, the relationship between content and concept was defined. Since each concept corresponds to one of the four contents, the content relationship for each concept was repeatedly inferred. Furthermore, as there are descriptions for each concept specified in korean textbooks, these were utilized to provide more detailed explanations to the VLM. The prompt (\hyperlink{step2.2}{Step 2.2}) to define the relationship between content and concept is provided at the bottom.

Finally, in Step 3, the results of Step 1 and Step 2 are synthesized and a concept is assigned to each question. However, since performance may be degraded if one of all concepts is directly selected, the concept candidate group was first filtered based on the result of Step 2. Specifically, both the question and the concept contain grade, and the content extracted in Step 2 was further utilized to restrict the list of the concepts that could correspond to the question. For example, if grade is 3 and the content is a '수와 연산', the candidate of the concept is reduced to 16. Through this candidate filtering, unnecessary search space can be reduced and performance degradation can be minimized. After that, the final concept was determined by using the question image and text, the filtered the concept list, and the description of each the concept as inputs together. The related prompt (\hyperlink{step3}{Step 3}) is provided at the bottom.

In Steps 2 and 3, since VLM can generate text different from the predefined content and concept, re-inference was performed up to 5 times when the generation result did not match the predefined texts for each features. Nevertheless, when matching results were not obtained, the procedure of mapping to the most similar value using edit distance \citep{levenshtein1966binary} was applied.

\clearpage

\begin{table*}[!t]
    \centering
    \hypertarget{step1}{}
    {\textbf{Step 1. Extracting question text}\par}
    \vspace{0.3em}
    \begin{tabularx}{\textwidth}{>{\raggedright\arraybackslash}p{0.15\textwidth} X}
        \hline
        System &
        당신은 한국의 수학 교육 전문가이자 고성능 OCR 시스템입니다. 이미지에 존재하는 모든 텍스트를 출력하는 것이 당신의 임무입니다.

        \vspace{0.5em}

        \begin{minipage}[t]{\linewidth}
        지시사항:
        \newline
        \noindent - 이미지에 존재하는 모든 텍스트만 출력하세요.
        
        \noindent - 이미지에 존재하는 텍스트 외의 내용은 출력하지 않습니다.
        \end{minipage}
        \vspace{0.001em}
        \\
        \hline
        User &
        다음 이미지는 한국의 초등학생 수학 문항입니다. 해당 이미지에서 OCR 을 통해 텍스트만 추출하려합니다.
        \newline
        [Question image]
        \\
        \hline
    \end{tabularx}
\end{table*}



        
        
        
        
        
        
        
        
        

\begin{table*}[!t]
\centering
\hypertarget{step2.1}{}
\textbf{Step 2.1. Assigning question content}\par
\vspace{0.3em}

\begin{tabularx}{\textwidth}
{>{\raggedright\arraybackslash}p{0.15\textwidth}
 >{\raggedright\arraybackslash}X}
\hline
System &
당신은 한국의 수학 교육 전문가이자 고성능 Classification 시스템입니다. 제공된 문항의 내용과 이미지를 분석하여 내용영역을 선택하는 것이 당신의 임무입니다.

\vspace{0.5em}

\begin{minipage}[t]{\linewidth}
지시사항:

\noindent - 문항을 자세하게 검토하여, 문항을 풀이할 때 요구되는 활동과 가장 일치하는 내용영역을 하나만 선택하세요.

\noindent - 내용영역은 다음과 같습니다.

\hspace*{1em}- 수와 연산: [수와 연산-description]

\hspace*{1em}- 변화와 관계: [변화와 관계-description]

\hspace*{1em}- 도형과 측정: [도형과 측정-description]

\hspace*{1em}- 자료와 가능성: [자료와 가능성-description]

\noindent - 출력은 반드시 4가지 내용영역 중 하나만을 선택해야 합니다.

\hspace*{1em}\texttt{수와 연산}
\end{minipage}
\tabularnewline
\hline
User &
다음 이미지는 한국의 초등학생 수학 문항입니다. 해당 이미지에서 OCR을 통해 텍스트만 추출하려 합니다.

[Question image]
\tabularnewline
\hline
\end{tabularx}
\end{table*}



        
        
        
        
        
        
        
        
        
        



\begin{table*}[!t]
    \centering
    \hypertarget{step2.2}{}
    {\textbf{Step 2.2. Content-concept relations prompt}\par}
    \vspace{0.3em}

    \begin{tabularx}{\textwidth}
        {>{\raggedright\arraybackslash}p{0.15\textwidth}
         >{\raggedright\arraybackslash}X}
        \hline

        System &
        당신은 한국의 수학 교육 전문가이자 고성능 Classification 시스템입니다.
        제공된 개념의 이름과 설명을 분석하여 내용영역을 선택하는 것이 당신의 임무입니다.

        \vspace{0.5em}

        \begin{minipage}[t]{\linewidth}
            지시사항:

            \noindent - 개념을 자세하게 검토하여, 개념과 가장 관련 있는 내용영역을 하나만 선택하세요.

            \noindent - 내용영역은 다음과 같습니다.

            \hspace*{1em}- 수와 연산: [수와 연산-description]

            \hspace*{1em}- 변화와 관계: [변화와 관계-description]

            \hspace*{1em}- 도형과 측정: [도형과 측정-description]

            \hspace*{1em}- 자료와 가능성: [자료와 가능성-description]

            \noindent - 출력은 반드시 4가지 내용영역 중 하나만을 선택해야 합니다.

            \hspace*{1em}\texttt{수와 연산}
        \end{minipage}
        \tabularnewline
        \hline

        User &
        다음은 한국의 초등학생 개념과 설명입니다.
        해당 개념에 적합한 내용영역을 선택하려 합니다.

        \noindent - 개념명: [concept name]

        \noindent - 개념 설명: [concept description]

        \vspace{0.5em}

        개념을 검토하여 4가지 내용영역 중 하나를 선택하세요.
        \tabularnewline
        \hline

    \end{tabularx}
\end{table*}



        
        
        
        




        

        
        
        
        



\begin{table*}[!t]
    \centering
    \hypertarget{step3}{}
    {\textbf{Step 3. Assigning question concept}\par}
    \vspace{0.3em}

    \begin{tabularx}{\textwidth}
        {>{\raggedright\arraybackslash}p{0.15\textwidth}
         >{\raggedright\arraybackslash}X}
        \hline

        System &
        당신은 한국의 수학 교육 전문가이자 고성능 Classification 시스템입니다.
        제공된 문항의 내용과 이미지를 분석하여 개념을 선택하는 것이 당신의 임무입니다.

        \vspace{0.5em}

        \begin{minipage}[t]{\linewidth}
            지시사항:

            \noindent - 문항을 자세하게 검토하여, 문항과 가장 일치하는 개념을 하나만 선택하세요.

            \noindent - 출력은 반드시 후보 개념 중 하나만을 선택해야 합니다.

            \hspace*{1em}\texttt{크기가 같은 분수 만들기}
        \end{minipage}
        \tabularnewline
        \hline

        User &
        다음은 한국의 초등학생 수학 문항입니다.
        해당 문항에 적합한 개념을 선택하려 합니다.

        \noindent - 수학 문항 이미지

        [Question image]

        \vspace{0.5em}

        \noindent - 수학 문제 및 해설 내용

        [Problem text]

        [Explanation text]

        \vspace{0.5em}

        \noindent - 후보 개념 목록:

        \hspace*{1em}- [개념명 1]: [개념 1-description]

        \hspace*{1em}- [개념명 2]: [개념 2-description]

        \hspace*{1em}- [개념명 3]: [개념 3-description]

        \hspace*{1em}- \ldots

        \hspace*{1em}- [개념명 N]: [개념 N-description]

        문항을 검토하여 후보 개념 중 하나를 선택하세요.
        \tabularnewline
        \hline

    \end{tabularx}
\end{table*}

\par\medskip

\raggedbottom
\FloatBarrier
\balance

\printcredits

\bibliographystyle{cas-model2-names}
\bibliography{cas-refs}





\end{document}